\documentclass[11pt,letterpaper]{article}
\usepackage[hyphens]{url} 
\usepackage{main}
\let\OldUrlFont\UrlFont \renewcommand{\UrlFont}{\small\OldUrlFont} 

\usepackage[hyphenbreaks]{breakurl}

\usepackage{xcolor}
\definecolor{darkblue}{rgb}{0, 0, 0.5}
\hypersetup{colorlinks=true,citecolor=darkblue, linkcolor=darkblue, urlcolor=darkblue}

\usepackage[utf8]{inputenc} 
\usepackage{microtype}
\usepackage{natbib}

\usepackage[tight-spacing = true]{siunitx} 

\usepackage{changepage}

\usepackage[aboveskip=1pt,labelfont=bf,labelsep=period,singlelinecheck=off]{caption}

\makeatletter
\renewcommand{\@biblabel}[1]{\quad#1.}
\makeatother

\title{\vspace{-15mm}Fast transcription of speech in low-resource languages\vspace{-10mm}}

\author{
Mark Hasegawa-Johnson\textsuperscript{\rm 1}, Camille Goudeseune\textsuperscript{\rm 1}, Gina-Anne Levow\textsuperscript{\rm 2}\\
\textsuperscript{\rm 1} University of Illinois \hspace{10mm}
\textsuperscript{\rm 2} University of Washington\\
\texttt{jhasegaw@illinois.edu}\vspace{-30mm}
}

\begin{document}
\maketitle

\begin{abstract}
We present software that, in only a few hours, transcribes
forty hours of recorded speech in a surprise language,
using only a few tens of megabytes of noisy text in that language,
and a zero-resource grapheme to phoneme (G2P) table.
A pretrained acoustic model maps acoustic features to phonemes;
a reversed G2P maps these to graphemes;
then a language model maps these to a most-likely
grapheme sequence, \textit{i.e.,} a transcription.
This software has worked successfully with corpora in
Arabic, Assam, Kinyarwanda, Russian, Sinhalese, Swahili, Tagalog, and Tamil.
\end{abstract}

\section{Fast transcription}

Designing and training automatic speech recognition (ASR) for a new
language $L$ normally requires weeks or years.
We consider the problem of how to do this within a few hours.
The name of our system, ASR24~\cite{asr24}, refers to the original task specification:
the ASR must be designed, trained,
and functioning within 24 hours of learning the identity of $L$,
using only data found on the public internet,
i.e., texts and recorded speech but not transcribed speech.
The timeline is
even tighter in practice, because it applies not just to
transcriptions produced by the ASR, but also to downstream applications such as machine
translation and named entity (NE) recognition, the wider context for this ASR problem.
To meet that deadline, one must design, train, and
run ASR within six hours of receiving the first $L$-text and $L$-speech data.

Such speed demands the reuse of components
that have previously been developed and optimized for other
languages.  Our system reuses published acoustic
models that have been trained in English, Hungarian, Russian, and
Czech, though because of lexical mapping constraints, the most
effective incident-language system only reuses the English acoustic
model.  The experimental variables considered here,
therefore, are then phone-to-word mapping methods and the language
modeling methods.  In ASR, phone-to-word mapping normally uses
a lexicon; we consider two other methods with better
computational efficiency but poorer transcription accuracy.
Several different language models (LMs) are considered, and the optimal
trade-off between computational and accuracy considerations is not clear.

Section~\ref{sec:methods} describes experimental methods.
Section~\ref{sec:evaluations} describes evaluation results.
Section~\ref{sec:conclusions} lists conclusions and ideas for future work.

\section{Methods}
\label{sec:methods}

Test systems are implemented using the Kaldi toolkit~\cite{kaldi}.
A Kaldi ASR has four weighted finite state transducer (wFST) components,
which are composed into a single wFST search graph:
\begin{enumerate}
\item The \textbf{H} (hidden Markov model) transducer's
  input symbols are pdf\/id's, Kaldi's name for senones~\cite{Hwang92} or
  triphone states~\cite{Lee90a}.  The edge
  weights (the probability of a senone given an acoustic frame) are
  computed by a neural net.  Details of the neural nets differ
  greatly among ASRs.
\item The \textbf{C} (context dependency) transducer converts
  senones to phones.  A senone is completely specified by the sequence of phones;
  the mapping from senone sequence to phone sequence is learned and implemented
  with a decision tree~\cite{Odell94}.
\item The \textbf{L} (lexicon) transducer maps phones to
  words.  This mapping is
  usually deterministic or nearly so.  It is implemented as a lookup
  table called a lexicon, because an ASR that tries to recognize all possible
  pronunciation variants fails because there are too many
  possibilities~\cite{Tajchman95}.  Because constructing this lexicon is one of
  the most labor-intensive steps in designing an ASR for a
  new language, doing this in just a few hours requires an unconventional approach.
\item The \textbf{G} (grammar) transducer computes the probability of a
  word sequence.  It is normally an LM trained from $L$-text.  In the typical ASR24 scenario,
  the text is harvested from uncurated online sources and is extremely noisy.
  Therefore, significant data cleaning may be needed.
\end{enumerate}

We now consider ASR24's H, C, L, and G transducers in turn.
Section~\ref{sec:AMs} describes the pre-trained H transducers,
also known as acoustic models (AMs).
Section~\ref{sec:context} describes how the C transducers map $L$-senones to $L$-phones.
Section~\ref{sec:lexicon} describes methods for mapping $L$-phones to $L$-words,
using tries, least common substrings, or lexicons.
Section~\ref{sec:cleaning} describes data cleaning for
improving the LM and vocabulary.
Section~\ref{sec:betterLMs} describes other improvements to the LM.

\subsection{Mapping acoustic frames to senones: acoustic models}
\label{sec:AMs}

Our acoustic models include open-source models in English,
Hungarian, Russian, and Czech.

The English AM was published as an extension of the
ASpIRE~\cite{aspire} chain model~\cite{chain}, part of the standard ASR toolkit Kaldi~\cite{kaldi2,kaldi}.
This extension was from conversational English
to a broader English vocabulary.  We hypothesized that it could extend to
other languages as well, and happily found that it did so, even to
languages with phone sets markedly different from English.

We also used three ASRs trained on eastern European
languages~\cite{phnrecgithub,phnrec}: Hungarian, Russian, and Czech.
These emitted SAMPA phones, which we converted to ASpIRE phones via table lookup.

\subsection{Mapping senones to phones: context dependency}
\label{sec:context}

H transducers in English, Hungarian, Russian, and Czech each generate
senones (pdf\/id's) in the corresponding language, but the phone-to-word
mappings considered in section~\ref{sec:lexicon} all use $L$-phones.
The C transducer maps senones to $L$-phones in two steps.
First, the C transducer distributed with each open-source acoustic model
maps senones to that language's phones.
Those are then mapped to $L$-phones by a
computational model of non-native speech perception (a ``mismatched
channel'' model, trained previously~\cite{HasegawaJohnson17,Jyothi2015aaai}).
The mismatch model computes the
probability $p(Y|X)$ that, say, an anglophone would
transcribe the English phone sequence $Y$ when listening
to the $L$-phone sequence $X$.
For computational reasons, we reduce the mismatch model to a
single best path, a many-to-one mapping between
$L$-phones and source language phones, implemented as a lookup table.

ASR24's design was motivated by speed.  Our previous crowdsourcing approaches,
PTgen~\cite{ptgengithub,ptgen} and PTgen with MCASR~\cite{mcasr},
transcribed 40 h in 3 to 5 days of elapsed time.
Although that was much faster than conventional ASR,
now we needed such transcriptions within 24 hours, and preferably within just a few hours.
Even thousands of crowdsourced workers cannot work that quickly,
because it takes a few days for word to spread among workers
that our task is interesting, pays well, and pays quickly.
So instead we aimed to run a collection of pretrained cross-language ASRs,
simulating a very small crowd of very fast crowdsourced workers.
(Because the working memory of these ``workers'' was much longer than that of humans,
now we could also skip the splitting of recordings into 1 s clips.)

Each ASR, such as one trained on professionally transcribed speech in half a dozen languages~\cite{sbs},
would emit $L$-phones, directly when possible, or by emitting words in its own trained language
which were then converted back into $L$-phones by that language's G2P
(such as a commercial Mandarin ASR~\cite{cvte}, whose AM
could not be used in isolation).

Then, as usual, the phone transcription from each ASR
would be wrapped up into a common format by MCASR,
whereupon the full set was aligned and coalesced by PTgen into
a single phone transcription (actually, a ``sausage'' wFST).

\subsection{Mapping phones to words: trie, LCS, and lexicon}
\label{sec:lexicon}

To convert an $L$-phone transcription
(or one of many, by stochastically traversing the sausage) to an $L$-word transcription,
we have tried three approaches.
The first approach reads a pronunciation dictionary into a trie
(a prefix tree of phones, whose leaves are the dictionary's words), and then greedily
matches input phones to the longest possible word.  Unsurprisingly, this is blazingly fast.
It is also easy to
tune and optimize.  For instance, we have adapted it to:
\begin{itemize}
    \vspace{-1mm}
  \setlength\itemsep{-1mm}
  \item prefer words that a downstream machine translator considers to be in-vocabulary
  \item simplify phones
  \item de-noise and sanitize input
  \item apply Soundex-style~\cite{soundex} soft matching
  \item choose, when several words match equally well (homonyms), the one that best approximates $L$'s statistics
  \item sometimes match shorter words, to approximate $L$'s
    measured word-length distribution.
\end{itemize}

The second approach is slower but more accurate, because when it matches phone strings
between a transcription and the pronunciation dictionary,
it uses the Longest Common Substring (LCS) algorithm
instead of the trie's greedy left-to-right method,
which for example in `Beethoven' mistakenly labels `th' a phone instead of stopping after the `t'.
This LCS approach works as follows.

For each word in the dictionary,
find the longest phone sequence that it has in common with the transcription.
Of the words whose pronunciations have a globally longest common (phone) substring,
if that substring approximates the word's entire pronunciation,
add that word to a list of candidates.
From that list, choose the word with the smallest substring-to-pronunciation Levenshtein distance,
add it to the word-transcription, and flag its corresponding phones as used.
Repeat this procedure until only noncontiguous phones remain.
(We have also experimented with choosing a word whose LCS is slightly shorter than
the global one but whose Levenshtein distance is considerably shorter.  We have also
applied the trie's tunings to this approach.)

With either approach, running multiple ASRs did reduce noise (i.e., the word error rate),
as one expects when averaging the readings of multiple sensors.
However, the phone-to-word conversion still introduced worse noise.
We guess that this noise is because the conversion to
phone strings, to an intermediate format between the input audio recording and the output word sequence,
loses information such as each phone's triphone context.
No matter how one tunes the phone-to-word converter, for instance by enlarging the
Soundex equivalence classes, it matches either too few words,
or far too many---a sizeable fraction of the entire lexicon---for an LM to prune back.
(The word error rate for Kinyarwanda was 103.2\% (106.3\% for LCS), for Sinhalese, 101.1\% (100.1\% for LCS):
considerably worse than the values in tables \ref{tab:sinhalese} and \ref{tab:kinyarwanda}.)
But such information is preserved when the phone representation stays within a wFST.
So we rejected these first two approaches in favor of a third:
a single lexicon mapping $L$-phones directly to $L$-words.

This third approach combines the AM with a pronunciation
lexicon and an LM, both of which are built from raw
text and a table of grapheme-to-phoneme (G2P) rules.  The combining is
done conventionally, by composing wFSTs, to make an ASR that is a single wFST.
The lexicon is created by applying prebuilt
G2P symbol tables~\cite{g2ps} to a list of all
unigrams (words) in the LM.  Training the LM
therefore becomes even more important than usual for the performance
of the end-to-end ASR.  The next two sections describe data cleaning
and algorithm improvements for the LM.

\subsection{Language modeling: Data cleaning}
\label{sec:cleaning}

The raw text is denoised with simple heuristics, and by discarding text
that contains graphemes foreign to the G2P, such as words using different alphabets.
A word-trigram LM is then built in a few minutes with standard SRILM tools~\cite{srilm,srilm2}.

In the meantime, we can manually improve the $L$-text with techniques appropriate to the particular text:
\begin{itemize}
\vspace{-1mm}
\setlength\itemsep{-1mm}
  \item add geographical place names from gazetteers
  \item remove word sequences that resemble Bible verses (common in low-resource languages, where a Bible translation is a significant fraction of all available text)
  \item extend the G2Ps to handle loanwords (also common in low-resource languages)
  \item support mixed case
  \item keep improving the gazetteers
  \item replace the naive trigram LM with more sophisticated LMs.
\end{itemize}
Every 2.5 h thereafter, fresh improved transcriptions can be produced.

\subsection{Better language models}
\label{sec:betterLMs}
Enhancing LMs may improve top-down information about expected outputs.  They may also improve recognition of  topic-relevant vocabulary that is poorly attested in the training data.  In particular, named entities (NEs) such as geographical locations have proven problematic for our ASR of low-resource languages.

To address this, we developed four strategies to systematically enhance our LMs, and ultimately ASR, for relevant NEs.
These techniques for creating NE-oriented class-based LMs incorporated a continuum of unsupervised and supervised class information. Although they focus on location NEs, they generalize to other types.
\begin{itemize}
    \vspace{-1mm}
  \setlength\itemsep{0mm}
  \item Unsupervised clustering with \textit{supervised expansion}: Word classes are created using unsupervised Brown clustering, based on a multi-threaded implementation \cite{Jaech-and-Ostendorf15}.  New NE terms, not attested in training text, are then added to the clusters with the highest density of NE terms.
  \item Unsupervised clustering with \textit{semi-supervised seeding}: Clusters are initialized with known NEs, before unsupervised Brown clustering.
  \item \textit{Supervised} classes: Classes are created for words in known NE classes.  Other words are treated as singleton clusters.
  \item NE-based \textit{data augmentation}: The LM's training corpus is augmented with NE-bearing sentences, namely translation-parallel $L$ sentences that correspond to English sentences containing NEs (found using a gazetteer and an off-the-shelf English NE recognizer~\cite{Finkel05}).
    Additional sentences are generated by stochastically replacing NEs in existing NE-bearing sentences with other ones.  The rate of data augmentation is tuned on a pair of development sets, one targeting NE-dense sentences, another targeting the overall corpus distribution.
\end{itemize}
For these unsupervised clustering methods, the number of clusters was varied between 100 and 1500. However, model quality, based on perplexity on a development set, was relatively insensitive to number of clusters.  Thus, the number of clusters for final models was set to 750.
Each of these class-based model variants was then interpolated with a word-based $n$-gram model with Kneser-Ney discounting.  NE classes were identified based on gazetteers or lexical match in Geo\-Names~\cite{geonames}.
The original corpus was also augmented with multiple duplications of gazetteer entries.

These new LMs were integrated with ASR24 through a tightly coupled model,
where a unigram list and class-based LM were directly composed into the ASR's wFST.

\section{Evaluations}
\label{sec:evaluations}

Downloading and uncompressing the archive files containing recorded speech typically took 30 minutes.
Acquiring and preparing the raw text took only a few minutes.

For real-time evaluations on a surprise language $L$,
we have been running ASR24 on a dedicated 56-core compute server.
Combining the AM, the pronunciation dictionary, and the LM into an ASR usually takes about 2 h.
This speed bottleneck may be due to the large size of the training text.
Once the ASR has been built, it takes only another 0.5 h to transcribe 40 h of speech.

Because geonames were important to those who were reading and translating
our transcriptions, to the $L$-language phrases in the raw text we appended
multiple copies of a gazetteer's $L$-language phrases.  (Even a dozen copies did not
risk overboosting or ``hallucinating'' geonames.)

Our primary goal is to make transcriptions within a few hours,
for which we resort to a trigram LM.
But to also make improved transcriptions within a few days,
ASR24 can use, instead of raw text and a G2P table,
an externally built more sophisticated LM (section~\ref{sec:betterLMs}).

ASR24 can run on a compute cluster via the usual \verb!qsub! mechanism,
but in practice the short run time offered by thousands
of CPU cores is outweighed by the delay between when the job is
submitted to the queue and when it finally starts,
a classic tortoise-and-hare scenario.
So it runs on a single 56-core compute server instead.

The speed bottleneck in ASR24 comes from its conventional wFST-based design.
This design's advantage is that composing wFSTs into a single wFST
produces a very quick ASR.  Furthermore, transcribing a few hundred speech recordings is
embarrassingly parallel, that is, ideally suited to a multicore compute server.
But the up-front cost of this is the composition.  The component wFSTs of some
LMs exceeded 6~GB; composing them would require terabytes of RAM and tens of hours.
Composition's complexity is the product of each wFST's arc count, which is intractably large
when each count already exceeds \num{e8}. 
Even multicore composition is inherently difficult~\cite{jw13}.
So ASR24 is practically limited to building a wFST no larger than 2~GB
(\num{3e7}~states, \num{e8}~arcs),
from an LM with \num{1.2e5}~words, \num{3e7}~bigrams, and \num{1.5e5}~trigrams.

\vspace{-2mm}
\subsection{Speed and word error rate}
\vspace{-1mm}

\paragraph{Sinhalese.} From a corpus of \num{6.7e5}~phrases with \num{8.6e6} words,
building the ASR took 67~minutes.  The ASR then transcribed 382~min of recorded speech
in 15~min, or \textbf{25$\times$ real time.}  The transcription had a \textbf{93.2\%~WER,}
when using a naive trigram LM augmented with geonames, stripped of Biblical phrases,
and with rudimentary sentence segmentation.

Improving the LM (section~\ref{sec:betterLMs}; good sentence segmentation)
yielded an ASR with the same speed, and a nearly identical \textbf{93.5\%}~WER.
Building this LM took 5~min, including parameter tuning based on dev sets.
Building the ASR from the LM took 40~min.

Five systems were implemented (table~\ref{tab:sinhalese}).
\begin{table}
  \caption{System descriptions and word error rate (WER) of ASR24 systems implemented in Sinhalese.
  A dash means that WER was not measured.\vspace{2mm}}
  \begin{tabular}{|l|p{2.18in}|l|}\hline
    & Description & WER \\ \hline
v1 &Trigram LM from raw text. &92.4\%\\
v2 &LM: culled Bible stopwords, \newline added gazetteer. &-\\
v3 & Class-based LM, unsupervised \newline clustering with supervised expansion. &-\\
v4 &v2, sentence segmenting. &93.2\%\\
v5 & Data augmentation-based LM, \newline with good sentence segmenting. &93.5\%\\\hline
  \end{tabular}
  \label{tab:sinhalese}
\end{table}

\paragraph{Kinyarwanda.} From \num{7.4e5}~phrases totalling \num{1.2e6}~words,
building the ASR took 76~min.  The ASR then transcribed 427~min of recorded speech
in 20~min, or \textbf{22$\times$} real time.  The transcription had a \textbf{87.1\%}~WER,
when using a naive trigram LM with mixed case, augmented with geonames from a gazetteer,
stripped of Biblical phrases, and with rudimentary sentence segmentation.

Improving the LM (section~\ref{sec:betterLMs};
better sentence segmentation, good truecasing) yielded an ASR
with the same speed, but a \textbf{95.0\%}~WER.
Building this LM took 5~min, including parameter tuning based on dev sets.
Building the ASR from this LM took 8~min. 

Nine systems were implemented (table~\ref{tab:kinyarwanda}).
Although the WER metric is simplistic for agglutinative languages like this,
it remains useful as a rough performance estimate.
\begin{table}
  \caption{System descriptions and WER of ASR24 systems implemented in Kinyarwanda.\vspace{2mm}}
  \begin{tabular}{|l|p{2.18in}|l|}\hline
    & Description & WER \\ \hline
v1 & Trigram LM from uppercase \newline raw text. & 88.1\%\\
v2 & LM: culled Bible stopwords, \newline added gazetteer.&-\\
v3 & v2, lowercased.&-\\
v4 & Class-based LM, unsupervised \newline clustering with supervised expansion &-\\
v5 & v2, mixed case, \newline larger gazetteer.&-\\
v6 & v5, truecased.&-\\
v7 & v5, sentence segmentation, \newline n' prefixes.&87.1\%\\
v8 & Class-based LM with supervised classes.&-\\
v9 & Data augmentation-based LM. \newline Good sentence segmentation, \newline n' prefixes, truecased.&95.0\%\\\hline
  \end{tabular}
  \label{tab:kinyarwanda}
\end{table}

\subsection{Variant spellings}

Because Kinyarwanda words often have variant spellings, we normalized spelling in both
the reference transcription and the ASR24-generated transcription before calculating WER.
We detected variants by mechanically normalizing spelling, and then applying unsupervised
clustering to that intermediate list of words.
Each cluster was then an equivalence class of variant spellings.

Mechanical normalizing of the distinct ``raw'' words in the two transcriptions consisted of
converting to lower case, removing accents, removing most punctuation,
and removing apostrophes except when in the middle of a word such as \textit{bw’indwara}.

To detect variant spellings in these normalized words,
we first excluded any ones shorter than 6 letters,
because such words are more likely to be truly distinct.
For the remaining 11,000 words, we calculated pairwise string edit distance
(\num{6.5e7} comparisons, taking 2 h with an off-the-shelf Levenshtein implementation).
Then we refined this distance measure by reweighting particular insertions, deletions,
and substitutions, reducing the unit cost for some operations (table~\ref{tab:recost}).
Others have used similar heuristic reweightings~\cite{katakana}.
(For speed, we didn't reweight distances of 3 or more.)

\begin{table}
  \caption{Non-unit costs for string-edit-distance operations for Kinyarwanda.\vspace{2mm}}
  \begin{tabular}{|l|l|}\hline
Cost & Operation \\ \hline
0.05 & Insert (or delete) an apostrophe \\
0.05 & Substitute Unicode apostrophes \\
0.1 & Insert a vowel between consonants \\
0.4 & Append a vowel \\
0.8 & Prepend a vowel, before a consonant \\
0.5 & Substitute one vowel for another \\
0.15 & Substitute `l' for `r' \\
0.3 & Substitute `m', `n' and `ng' \\\hline
  \end{tabular}
  \label{tab:recost}
\end{table}
For example, \textit{bwindwara} and \textit{bw’indwara} had a conventional distance of 1.0, reweighted to 0.05 with the apostrophe rule;
\textit{ingengabitekerezo} and \textit{ngengabitekerezo} reweighted from 1.0 to 0.8 with the prepended vowel rule.
(We also tried dividing this distance by the word pair's mean word length,
but that made clustering worse.)
Using this set of pairwise distances, we then built, for each word $w$,
an array $A(w)$ of (distance, word) pairs, sorted by distance, for quick lookup during clustering.

We grew an initial cluster $C$ from each word $w$,
by continually adding the word $v \in A(w)$ nearest to $w$,
as long as $v$ was within a threshold distance (1.5) of each word in $C$ so far.
From these initial clusters, we discarded singletons, and also discarded any one that was a subset of another.
When a word ended up in more than one cluster,
we kept it in only the cluster whose centroid was closest to it.
We estimated a cluster's centroid only as needed, as the word whose median distance to the other words was minimal.

We might have improved the recognition by considering the words' context, like an implicit LM.
Similar contexts for two similarly spelled words argue for them really being variant spellings~\cite{katakana}.
But as it was, this method already turned out to be better suited to detecting variant spelling
than either hierarchical agglomerative clustering or complete linkage clustering.

\section{Conclusions and future directions}
\label{sec:conclusions}

We have described ASR24, a system that designs an ASR for a
surprise language $L$ within 2 or 3 hours, using zero
transcribed audio in $L$.  The system consists of
re-used acoustic models, mapped to $L$-phones using a
mismatched crowdsourcing channel model, mapped to words using
published grapheme-to-phoneme symbol tables, and then finally
regularized using a trained LM for $L$.

Future work will consider the problem of morphological complexity.
Agglutinative languages like Kinyarwanda and Tamil can have intractably large word lists,
due to their affixes' combinatorial explosion.
For such languages, instead of working with full words,
it may be better to work directly with stems and affixes,
found automatically by a tool like Morfessor~\cite{morfessor}.

\vspace{4mm}
\section*{Acknowledgment}
This work was funded by the DARPA program ``Low resource languages for emergent incidents (LORELEI),'' DARPA-BAA-15-04.
\vspace{4mm}

\bibliography{main}
\bibliographystyle{abbrv}
\end{document}